%% file: root.tex
\documentclass[letterpaper, 10 pt, conference]{ieeeconf}  

\IEEEoverridecommandlockouts                              

\makeatletter
\let\NAT@parse\undefined
\makeatother

\usepackage{graphics} 
\makeatletter
\let\MYcaption\@makecaption
\makeatother
\usepackage[font=footnotesize]{subcaption}
\makeatletter
\let\@makecaption\MYcaption
\makeatother
\usepackage{graphicx}
\usepackage[activate=false]{microtype}
\usepackage{amsmath} 
\usepackage{amssymb}  
\usepackage{bm}
\newcommand{\bs}[1]{\boldsymbol{#1}}
\usepackage{url}

\usepackage[noend]{algpseudocode}
\usepackage{algorithm}
\usepackage{balance}
\usepackage{pifont} 
\usepackage[square, comma, numbers,sort]{natbib}
\usepackage{multirow}
\usepackage[table,dvipsnames]{xcolor}
\usepackage[pagebackref=true,breaklinks=true,colorlinks,bookmarks=false]{hyperref}

\hypersetup{
linkcolor=BrickRed
,citecolor=Green
,filecolor=Mulberry
,urlcolor=NavyBlue
,menucolor=BrickRed
,runcolor=Mulberry
,linkbordercolor=BrickRed
,citebordercolor=Green
,filebordercolor=Mulberry
,urlbordercolor=NavyBlue
,menubordercolor=BrickRed
,runbordercolor=Mulberry
}

\newcommand{\acronym}{FusionSense}

\def\etal{\textit{et al. }}

\input{sections/00-title_author}

\begin{document}

\maketitle
\thispagestyle{empty}
\pagestyle{empty}


\input{sections/0-Abstract}

\input{sections/1-Introduction}

\input{sections/2-Related}

\input{sections/3-Method}

\input{sections/4-Experiment}

\input{sections/5-Conclusion}


\clearpage
{\small
\bibliographystyle{IEEEtranN}
\balance
\bibliography{root}
}
\newpage
\section*{Appendix}
\label{sec:appendix}
\input{sections/6-Appendix}
\end{document}

%% file: sections/00-title_author.tex
\title{\LARGE \bf
FusionSense: Bridging Common Sense, Vision, and Touch \\for Robust Sparse-View Reconstruction
}

\author{Irving Fang$^{1,*}$, Kairui Shi$^{1,*}$, Xujin He$^{1, *}$, Siqi Tan$^{1}$, Yifan Wang$^{1}$, Hanwen Zhao$^{1}$, Hung-Jui Huang$^{2}$, \\ Wenzhen Yuan$^{3}$, Chen Feng\textsuperscript{1,\ding{41}}, Jing Zhang\textsuperscript{1,\ding{41}}\\
{\tt\small \url{https://ai4ce.github.io/FusionSense/}}
\thanks{$^{1}$New York University, Brooklyn, NY 11201, USA}
\thanks{$^{2}$Carnegie Mellon University, Pittsburgh, PA 15289, USA}
\thanks{$^{3}$University of Illinois, Urbana-Champaign, Champaign, IL 60606, USA}
\thanks{$^{*}$Equal contributions.}
\thanks{\ding{41} Corresponding authors {\tt\small \{\href{mailto:cfeng@nyu.edu}{cfeng},\href{mailto:jz6676@nyu.edu}{jz6676}\}@nyu.edu}. The work was supported in part through NSF grants 2024882, 2152565, and 2238968.}
}

\IEEEaftertitletext{
\begin{center}
    \vspace{-7mm}
    \centering
    \captionsetup{type=figure}
    \includegraphics[width=\textwidth]{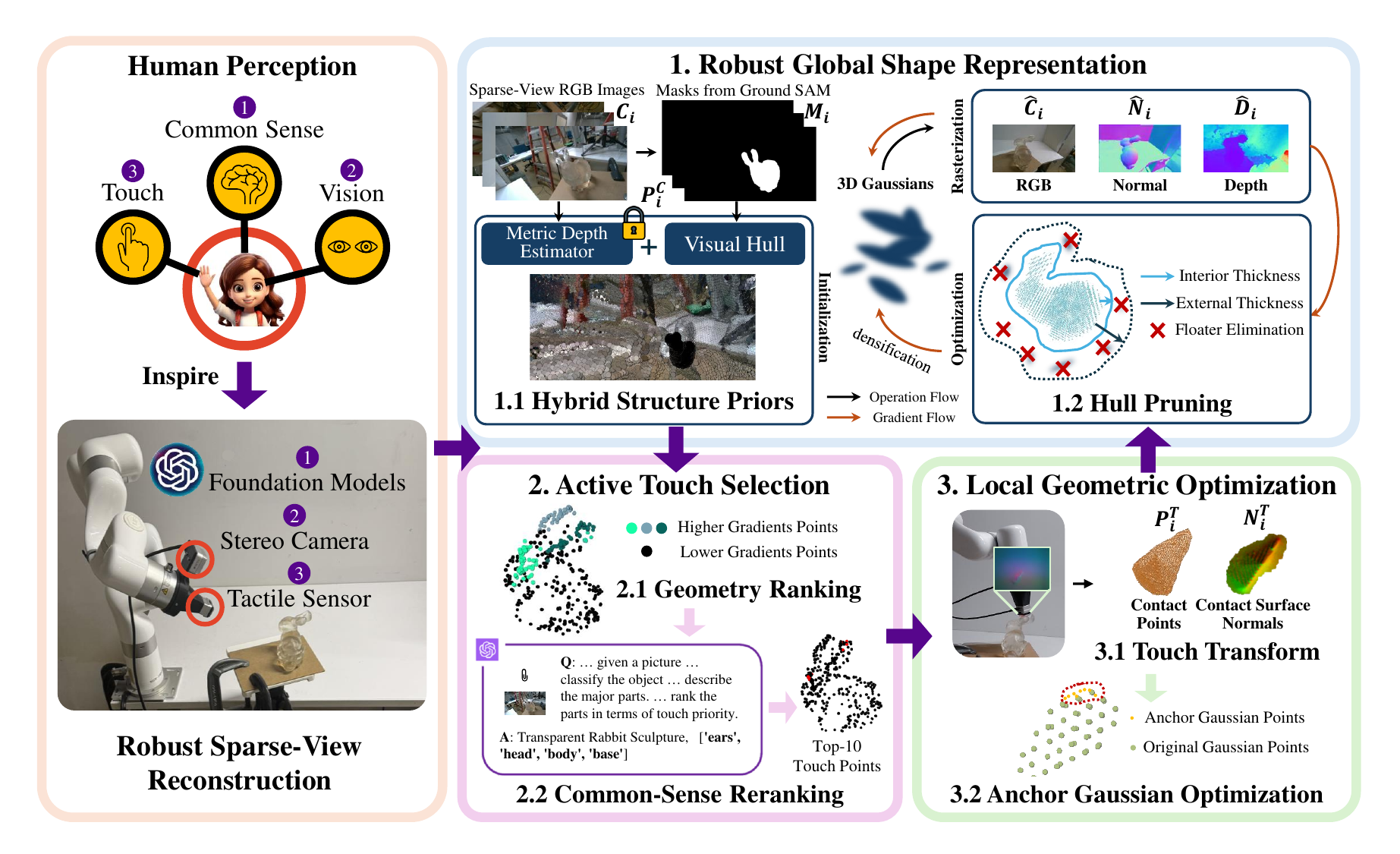}
    \vspace{-7mm}
    \captionof{figure}{\textbf{Overview of \acronym.} Inspired by human perception, \textbf{\acronym} integrates common sense from foundation models with \textit{sparse-view} data from both vision and touch through 3D Gaussian Splatting, enabling efficient and robust 3D reconstruction of a robot's surroundings. Our proposed system features three core modules: (i) robust global shape representation, (ii) active touch point selection on the object, and (iii) local geometric optimization. }
    \label{fig:fig1}
\end{center}%
}

%% file: sections/0-Abstract.tex
\begin{abstract}

Humans effortlessly integrate common-sense knowledge with sensory input from vision and touch to understand their surroundings. Emulating this capability, we introduce \textit{FusionSense}, a novel 3D reconstruction framework that enables robots to fuse priors from foundation models with highly sparse observations from vision and tactile sensors. FusionSense addresses three key challenges: (i) How can robots efficiently acquire robust global shape information about the surrounding scene and objects? (ii) How can robots strategically select touch points on the object using geometric and common-sense priors? (iii) How can partial observations such as tactile signals improve the overall representation of the object? Our framework employs 3D Gaussian Splatting as a core representation and incorporates a hierarchical optimization strategy involving global structure construction, object visual hull pruning and local geometric constraints. This advancement results in fast and robust perception in environments with traditionally challenging objects that are transparent, reflective, or dark, enabling more downstream manipulation or navigation tasks. Experiments on real-world data suggest that our framework outperforms previously state-of-the-art sparse-view methods. All code and data are open-sourced on the project website.
\end{abstract}

%% file: sections/1-Introduction.tex
\section{Introduction}

Humans exhibit an extraordinary ability to perceive their surroundings by seamlessly integrating common-sense knowledge, vision, and touch, even when presented with sparse or incomplete views \cite{hutmacher2019there}. Common-sense reasoning helps bridge gaps in sensory data, vision offers a broad understanding of the environment, and touch provides fine-grained information about texture and material properties through direct physical interaction \cite{helbig2007optimal}. This synergy between cognitive and sensory inputs inspires more intuitive and efficient robotic perception in complex environments \cite{allen1984surface, navarro2023visuo}.

Despite recent advances, current robotic perception systems have yet to fully harness the multimodal capabilities that humans naturally employ. Emerging techniques like 3D Gaussian Splatting (3DGS) \cite{kerbl20233d} show potential for flexible and efficient 3D reconstruction of intricate structures. However, vision-based methods \cite{yu2024gaussian, wang2024dust3r}, especially those dependent on \textit{sparse-view observations} \cite{yang2024gaussianobject}, continue to face challenges such as occlusion, suboptimal lighting conditions, and difficult surfaces like transparent \cite{cai2024neuralto}, reflective \cite{liu2023nero}, or dark objects \cite{swann2024touch}. Approaches such as \cite{comi2024touchsdf} leverage pre-trained models like DeepSDF \cite{park2019deepsdf} for shape completion, yet they still struggle with objects possessing unique geometries or intricate details. Conversely, high-resolution optical tactile sensors \cite{yuan2017gelsight, wang2021gelsight} can overcome these limitations through direct physical interaction with high-resolution sensing, yet they have a limited sensing range. For example, the reinforcement learning strategy in \cite{shahidzadeh2024actexplore} takes a cobot 1,631 touches to fully explore the surface of a banana in the YCB dataset \cite{calli2015ycb}, which has a surface area of only 216 cm$^\text{2}$. Furthermore, while multimodal methods combining visual and tactile data have shown promise for improving object perception and 3D reconstruction, passive touch strategies often significantly increase the number of actions needed \cite{suresh2022shapemap, swann2024touch}.

To overcome these limitations, we present \textbf{\acronym{}}, a novel 3D reconstruction framework that integrates priors from foundation models with sparse observations from both vision and tactile sensors. At the core of our framework is 3D Gaussian Splatting, which provides an efficient and scalable means to represent the environment. In this framework, surface normal supervision is highlighted to enrich both global and local geometric details \cite{bae2024rethinking, turkulainen2024dn, asuperprimitive, cao2024supernormal}. Specifically, \acronym{} is built upon three key modules: (\textbf{i}) \textbf{Robust Global Shape Representation}, where hybrid structure priors are introduced to initialize geometry and ensure multi-view consistency alongside a hull-pruning constraint to guide optimization for both the scene and the object; (\textbf{ii}) \textbf{Active Touch Selection}, based on the observation that high gradients in 3DGS represent complex structures or mismatches between splatting and the image, combined with common-sense knowledge from foundation models for decision-making; and (\textbf{iii}) \textbf{Local Geometric Optimization}, where new anchor Gaussians are added to guide fine detail optimization, with geometric normals supervised by the high-resolution tactile feedback provided by the GelSight sensor.

These innovations lead to the following key contributions:

\begin{enumerate} 
\item We propose a novel 3D reconstruction framework for scenes and objects that fuses priors from foundation models with \textit{sparse observations from visual and tactile sensors}, exploiting the unique strengths of each modality. We also develop an active touch strategy driven by geometric and common-sense cues, enhancing perceptual granularity with fewer robot actions. This framework can handle objects that are traditionally challenging for 3D reconstruction, such as transparent, reflective, or dark objects.
\item We propose a novel hierarchical optimization strategy designed for 3DGS. This strategy incorporates object hull pruning to guide the optimization process and introduces anchor Gaussians at the local level, supervised by surface normals captured from tactile signals, to refine fine-grained details. Our work is the first to natively incorporate tactile signals into 3DGS. 
\item We deploy our algorithm on a real robot, demonstrating its competitive ability to reconstruct surroundings with challenging objects under highly sparse observations.
\end{enumerate}

%% file: sections/2-Related.tex
\section{Related Work}
\subsection{Visuo-Tactile Robot Perception}
Roboticists have long been exploring tactile sensing. In 1984, Bajcsy and Goldberg \cite{Goldberg1984ActiveTA} had already explored tactile surface reconstruction with primitive tactile sensors. Recently leap in tactile technology \cite{yuan2017gelsight, uskin} enabled great progress in object classification \cite{corcodel2020interactive, 2014Schmitz}, deformable object manipulation \cite{Burns2022, Kaboli2016, she2021cable}, industrial insertion \cite{2019DOng,Dong2021}, etc. 

\textbf{In tactile-only reconstruction}, researchers often employ an active strategy to select touch points due to the limited coverage area of tactile sensors. Many of them \cite{Yi2016, Jamali2016} chose the Gaussian Process Implicit Surface as the representation for the shape, of which the derived uncertainty drives the selection strategy. Matsubara \etal \cite{MATSUBARA2017314} used end-effector travel distance as another constraint to accelerate the procedure, while Shahidzadeh \etal \cite{shahidzadeh2024actexplore} sidestepped Gaussian Process and utilized reinforcement learning for an exploration policy. 

\textbf{In visuo-tactile works}, visual signals can provide a rough global shape of the object, greatly reducing the number of touches and enabling passive touch strategies. Swann \etal \cite{swann2024touch} and Suresh \etal \cite{suresh2022shapemap} employed a grid-like, exhaustive touch strategy. Smith \etal \cite{smith20203d} only considered touch patch at the grasping spot. For active touch, Smith \etal \cite{smith2021active} learned a strategy in simulation, while Bj{\"o}rkman \etal \cite{bjorkman2013enhancing} and Wang \etal \cite{wang20183d} again employed uncertainty in Gaussian Process. A key observation is that the uncertainty in the Gaussian Process usually comes from a lack of visual signal on certain parts. The part itself may be otherwise unremarkable. However, our active strategy also focuses on the geometrically complicated and fine-grained parts. In addition, our method is the first to employ the state-of-the-art Gaussian Splatting \cite{kerbl20233d} method instead of a simple baseline method for initial reconstruction, the first to employ multiple foundation models, and the first to efficiently fuse tactile signal natively into Gaussian Splatting \cite{kerbl20233d}, unlike in Touch-GS \cite{swann2024touch} where the tactile signal is still incorporated via Gaussian Process Implicit Surface~\cite{williams2006gaussian}.

\subsection{Gaussian Splatting for 3D Reconstruction}

Gaussian Splatting \cite{kerbl20233d} is a fast and efficient method for 3D reconstruction and radiance field rendering, representing scenes with Gaussian primitives to preserve continuous volumetric properties while enabling rapid optimization and real-time rendering. DN-Splatter \cite{turkulainen2024dn} improves upon this by introducing geometric normal supervision, enhancing geometric accuracy, particularly in textureless regions, but its performance is limited under sparse-view observations. GaussianObject \cite{yang2024gaussianobject} is designed for object reconstruction in sparse-view settings and uses the visual hull to initialize the object point cloud, though its optimization process differs from ours. TouchGS \cite{swann2024touch} optimizes 3DGS using GPIS results~\cite{williams2006gaussian} derived from dense tactile observations (e.g., 632 touches), making it inefficient for robotic applications. Inspired by these methods, we present the first framework to fuse common sense and sparse observations from both vision and touch using Gaussian primitives without being limited by the number of touches.

%% file: sections/3-Method.tex
\section{Method}\label{sec:method}
\subsection{Problem Formulation}

Our goal is to represent a previously unseen scene, $S$, using a set of differentiable 3D Gaussian primitives, $\mathcal{G} = \{ \bs{G}_k: \bs{p}_k, \bs{q}_k, \bs{s}_k, o_k, \bs{c}_k \}_{k=1}^{K}$. The geometry of each Gaussian $\bs{G}_k$ is parameterized by its center $\bs{p}_k \in \mathbb{R}^{3}$, rotation quaternion $\bs{q}_k\in\mathbb{R}^{3}$, and scaling vector $\bs{s}_k\in\mathbb{R}^{3}$. The appearance parameters include opacity $o_k\in \mathbb{R}$, and color $\bs{c}_k\in\mathbb{R}^{3}$. Rendering a new view is achieved by projecting 3D Gaussians into 2D space from the camera's perspective. The resulting 2D Gaussians are depth-sorted globally and then alpha-composited using the discrete volume rendering equation to compute the final pixel colors, $\hat{C}$, depth estimation, $\hat{D}$, and Normal estimation, $\hat{N}$ \cite{turkulainen2024dn}:
\begin{equation}
\hat{C} = \sum_{k \in N} c_k \alpha_k T_k, \hat{D} = \sum_{k \in N} d_k \alpha_k T_k, \hat{N} = \sum_{k \in N} n_k \alpha_k T_k,
\end{equation}
where $T_k = \prod_{j=1}^{k-1} (1 - \alpha_j)$ is the accumulated transmittance at pixel location $p$ and $\alpha_k$ is the blending coefficient for a Gaussian with center $\mu_k$ in screen space:
\begin{equation}
\alpha_k = o_k \cdot \exp \left( -\frac{1}{2} (p - \mu_k)^\text{T} \Sigma_k^{-1} (p - \mu_k) \right).
\end{equation}

In particular, we are interested in a challenging object $O$ that may be transparent, reflective, or dark. We aim to reconstruct it in $O_\text{rec}$ (in this case, 3D Gaussian primitives) as close to the original object as possible.

To this end, we collect the following sparse observations from vision and tactile sensors and will fuse them with priors from foundation models: 

\begin{itemize}
    \item \textbf{Color} $\bs{C}_i$ and \textbf{depth} $\bs{D}_i$ \textbf{images}, and their \textbf{pose} $\bs{P}^\text{C}_i $ in the world frame with the following dimensions:
    $$\bs{C}_i \in \mathbb{R}^{1280\times 720\times 3},~\bs{D}_i \in \mathbb{R}^{1280\times 720\times 1},~\bs{P}^\text{C}_i \in SE(3).$$
    \item \textbf{Tactile} signal $\bs{T}_i$ and its \textbf{pose} $\bs{P}^\text{T}_i$ in the world frame with the following dimension:
    $$\bs{T}_i \in \mathbb{R}^{240\times 320\times 3},~\bs{P}^\text{T}_i \in SE(3).$$
\end{itemize}

Note that tactile signals are saved as RGB images to be processed later. Also, note that the color and depth images are aligned so they share the same pose.

\subsection{Method Overview}
Our method can be divided into three modules:
\subsubsection{Robust Global Shape Representation} This module leverages hybrid geometry priors and object hull pruning to optimize a global 3D representation, denoted as $\mathcal{G}$, that contains the scene and the object of interest $\bs{O}$. The hybrid geometry prior combines monocular depth estimates $\bs{\overline{D}}_i$ \cite{hu2024metric3d}, camera poses $\bs{P}^\text{C}_i$, and visual hull results $\bs{\overline{O}}$ \cite{laurentini1994visual} to produce an initial representation $\mathcal{G}'$. During optimization, hull pruning eliminates floating artifacts and ensures a clean representation of the initial reconstructed object $\bs{O}'$, derived from both $\bs{\overline{O}}$ and the global shape $\mathcal{G}$. $\mathcal{G}'$ is supervised with color $\bs{C}_i$, depth $\bs{D}_i$, and normal priors $\bs{\overline{N}}_i$ from \cite{bae2024rethinking}.
\subsubsection{Active Touch Selection} This module proposes touch points $\bs{t}_i$ on $\bs{O}'$ where tactile feedback is needed. The robot then collects tactile signals $\bs{T_i}$ at these points. It consists of two sub-modules: 
    \begin{itemize}
        \item A geometry-focused module that ranks points in regions with high gradients in 3DGS, indicating intricate structures or discrepancies between splatting and the image.
        \item A common-sense-driven module that utilizes large vision and language models (VLMs) to rerank points from the previous module, integrating common-sense knowledge from VLMs to enhance decision-making.
    \end{itemize}
\subsubsection{Local Geometric Optimization} This module takes in $\bs{T}_i$ and the contact masks $\bs{M}_i^\text{T}$, surface normals $\bs{N}_i^\text{T}$, and contact points $\bs{P}_i^\text{T}$ \cite{suresh2022shapemap}, introducing an anchor Gaussian optimization strategy. Anchor Gaussians $\bs{G}^\text{T}$ are initialized from $\bs{P}_i^\text{T}$ and further refined using $\bs{N}_i^\text{T}$ and global context. By integrating tactile signals $\bs{T}_i$ into the global representation $\mathcal{G}$, this module refines local geometric details. 

\subsection{Robust Global Shape Representation}
\label{subsec: global}
\begin{figure}
    \vspace{1mm}
    \begin{center}
    \includegraphics[width=\columnwidth]{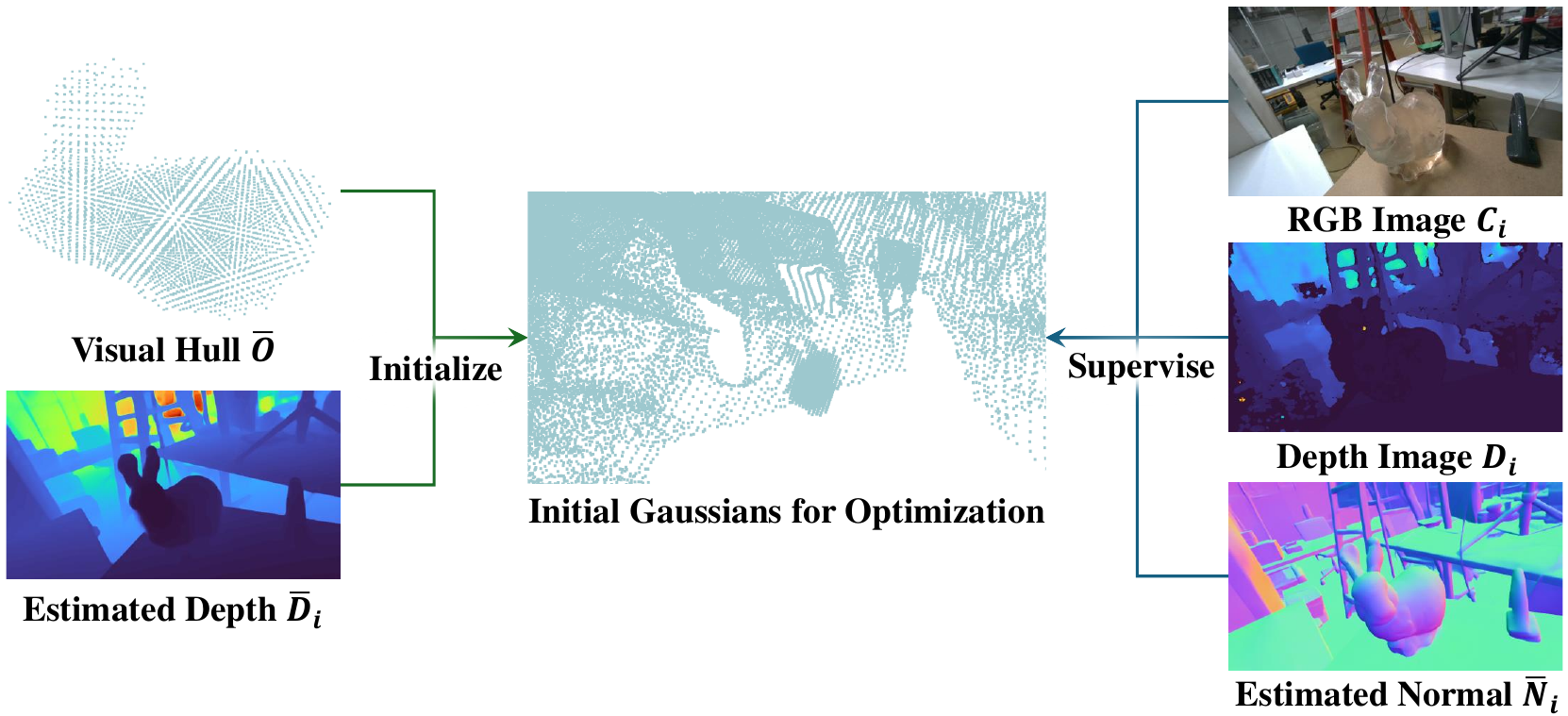}
    \vspace{-5mm}
    \caption{We use visual hull and estimated depth to initial our Gaussians and use RGB, depth, and estimated normal to supervise the training.}
    \label{fig:global shape}
    \end{center}
    \vspace{-7mm}
\end{figure}

To obtain a robust global representation $\mathcal{G}$, we introduce hybrid structure priors and hull pruning strategies. Specifically, we adopt a variant of 3DGS \cite{turkulainen2024dn} that incorporates surface normal supervision. 

Hybrid structure priors are employed to ensure multi-view consistency. First, we estimate the coarse geometry $\bs{\overline{O}}$ of the target object using a visual hull \cite{laurentini1994visual}, which is constructed by combining camera poses $\bs{P}^\text{C}_i$ and segmented silhouettes $\bs{M}_\text{o}$ extracted via Grounded SAM 2 \cite{ren2024grounded}. This method is independent of surface appearance, resilient to challenging materials, and key to our success with objects that are otherwise challenging to traditional reconstruction methods. Next, we acquire the surrounding coarse geometry $\bs{\overline{S}}$ using monocular depth priors $\bs{\overline{D}}_i$ from depth foundation model Metric3D v2 \cite{hu2024metric3d}, along with the corresponding camera poses $\bs{P}^\text{C}_i$. These hybrid structure priors are fused by applying distance thresholds $\bs{\tau}_\text{d}$ to integrate $\bs{\overline{O}}$ and $\bs{\overline{S}}$ to produce the initial global representation $\mathcal{G}'$, which contains the initial reconstructed $\bs{O}'$, for further optimization.

During the subsequent optimization process, we design hull pruning to remove the floaters in the exterior region outside the hull $\bs{\overline{O}}$. Gaussian primitives are particularly sensitive to these floaters, as they can slow convergence and result in suboptimal outcomes, especially when dealing with sparse observations. Hull pruning is achieved by introducing a thin shell $\bs{\overline{O}}_\text{s}$ surrounding the hull $\bs{\overline{O}}$. $\bs{\overline{O}}_\text{s}$ is defined by two thickness parameters: an interior thickness $t_\text{s}^\text{i}$ and an external thickness $t_\text{s}^\text{e}$. In our setup, $t_\text{s}^\text{i}$ is set to be larger than 5 mm, corresponding to the voxel grid resolution of the visual hull, while $t_\text{s}^\text{e}$ is empirically set to 2 cm. Then, similar to \cite{turkulainen2024dn}, we utilize RGB $\bs{C}_i$ and depth $\bs{D}_i$ image and normal $\bar{N}_i$ estimated from normal foundation model DSINE \cite{Dsine} to respectively supervise $\hat{C}, \hat{D}, \hat{N}$ from our $\mathcal{G}'$, as can be seen in Fig. \ref{fig:global shape}.

\subsection{Active Touch Selection}

An active touch strategy with geometric and common-sense cues can reduce the number of touches needed.

\subsubsection{Geometry} \label{subsubsec: geometry}

We capitalize on the design of the original 3DGS \cite{kerbl20233d} that high gradients at each Gaussian primitive indicate rapid changes in spatial features and larger discrepancies between the rendering generated by the Gaussians and the image, which means a need for further optimization.

Given the objective $\bs{O}'$ for tactile interaction, we use the densification mean value from Sec. \ref{subsec: global} as the gradient threshold $\tau_g$ to select some Gaussian primitives $\bs{G}_k'$. Next, we apply DBSCAN \cite{DBSCAN} algorithm to cluster $\bs{G}_k'$, filtering out outliers. Then, all the selected Gaussians are ranked based on mean gradient values in its cluster, forming a ranking $\mathcal{R}_G$.

\subsubsection{Common Sense} The sub-module provides another ranking $\mathcal{R}_C$ by leveraging common-sense knowledge from vision-language models (VLMs). 

First, we randomly select one color image from the captured images $\bs{C}_i$ and prompt GPT-4-o \cite{openai2023gpt} with the image and descriptive text to obtain a classification label and a list of relevant part names, along with a ranking $\mathcal{R}_p$ of the parts based on priority in touching.

\begin{figure}
\begin{center}
\includegraphics[width=0.9\columnwidth]{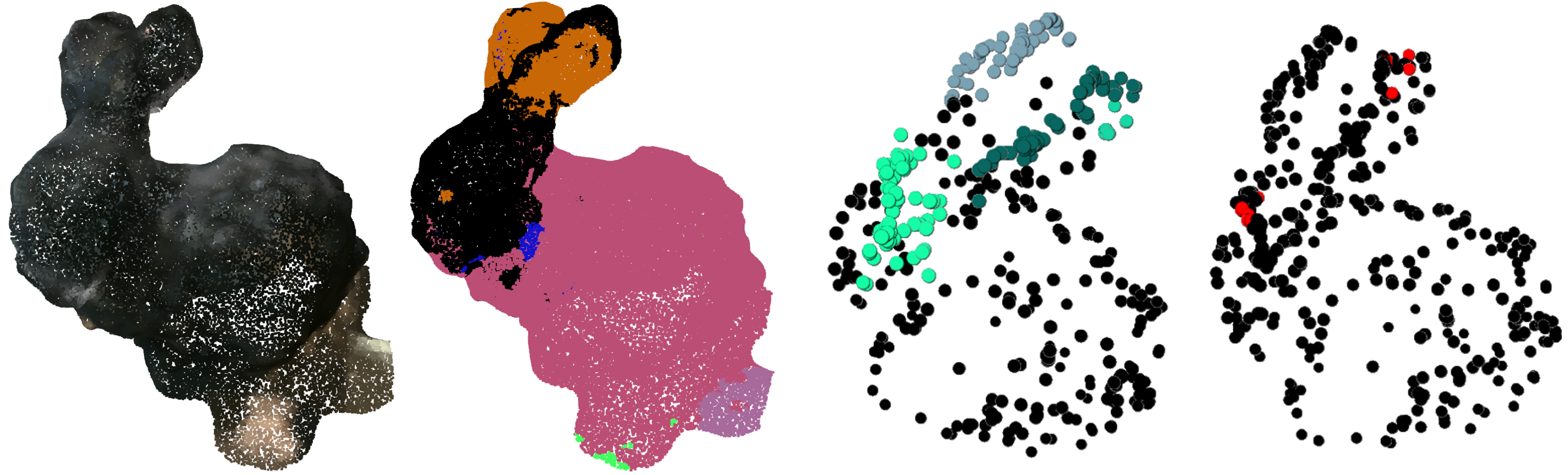}
\caption{\textbf{(1)} Point Cloud Extracted from $\bs{O}'$. \textbf{(2)} Part Segmentation from PartSLIP. \textbf{(3)} High Gradient Gaussians. \textbf{(4)} 10 Selected Touch Points $\bs{t}_i$.}
\label{fig:active touch}
\end{center}
\vspace{-5mm}
\end{figure}

To ground this common-sense ranking $\mathcal{R}_p$ to the object $\bs{O}'$, we utilize a zero-shot open-vocabulary part segmentation model, PartSLIP \cite{liu2023partslip}. Based on a textual prompt of parts names, PartSLIP classifies each point of an extracted point cloud from $\bs{O}'$ into a specific part as in Fig. \ref{fig:active touch}. From Sec. \ref{subsec: global}, we know the coordinates of every Gaussian in $\bs{G}_k'$ and every point in the extracted point cloud from $\bs{O}'$ as can be seen in Fig. \ref{fig:active touch}. We also know each point's ranking in $\mathcal{R}_p$. So, we iterate through $\bs{G}_k'$, assigning every Gaussian a rank based on the closest point in the point cloud, thus forming another ranking $\mathcal{R}_C$ for selected Gaussians $\bs{G}_k'$.

We then sort $\bs{G}_k'$ based on $\mathcal{R}_C$ first and then $\mathcal{R}_G$, ensuring that even if PartSLIP in the second sub-module fails to work properly, we still have a reasonable, geometrically sensible touch sequence $\bs{t}_i$, as seen in Fig. \ref{fig:active touch}

\subsection{Local Geometric Optimization} 

This module enhances local geometric detail by transforming the tactile signal $\bs{T}_i$ into contact masks $\bs{M}_i^\text{T}$, surface normals $\bs{N}_i^\text{T}$, and contact points $\bs{P}_i^\text{T}$~\cite{suresh2022shapemap}. We then introduce anchor Gaussian optimization to integrate the tactile signals $\bs{T}_i$ into the global representation $\mathcal{G}$.

Given a tactile signal $\bs{T}_i$ as an RGB image, because the tactile sensor is made of a gel patch that has consistent optical properties across all its surface, we can calculate a mapping between surface gradients $(\frac{\partial f}{\partial x}, \frac{\partial f}{\partial y})$ and the RGB value at a given location $(r, g, b, x, y)$ based on photometric stereo \cite{Woodham1992}. In practice, this mapping is acquired by pressing a ball with a known radius on the gel patch and recording the corresponding tactile image. Then, a multi-layer perceptron can be trained after manually labeling the deformation caused by the ball. Assuming that the gel patch is the zero level surface of a scalar field $f(x, y) - z$, the contact surface normal $\bs{N}_i^\text{T}$ can be derived from the surface gradient as $(\frac{\partial f}{\partial x}, \frac{\partial f}{\partial y}, -1)$ \cite{yuan2017gelsight}. Applying a Poisson solver to integrate the surface gradients gives us a depth map of the gel patch's shape. We can then acquire a contact mask $\bs{M}_i^\text{T}$ and, consequentially, contact points $\bs{P}_i^\text{T}$ with a depth threshold.

Contact points $\bs{P}_i^\text{T}$ are added as anchor Gaussians $\bs{G}^\text{T}$ due to the scale difference between $\bs{T}_i$ and visual signals $\bs{C}_i$. Treating $\bs{P}_i^\text{T}$ as ground truth, we fix the center $\bs{p}^\text{T}$ and opacity $o^\text{T}$ of $\bs{G}^\text{T}$, while optimizing the rotation $\bs{q}^\text{T}$, scale $\bs{s}^\text{T}$, and color $\bs{c}^\text{T}$. Notably, we apply Gaussian normal supervision directly to $\bs{G}^\text{T}$ instead of normal images. This allows the integration of $\bs{G}^\text{T}$ into $\mathcal{G}$, combining local surface normals $\bs{N}_i^\text{T}$ with global information $\bs{C}_i$, $\bs{D}_i$ to refine the final geometry.

\vspace{-3mm}


%% file: sections/4-Experiment.tex
\section{Experiment}

\begin{figure*}
    \vspace{1mm}
    \begin{center}
    \includegraphics[width=0.9\textwidth]{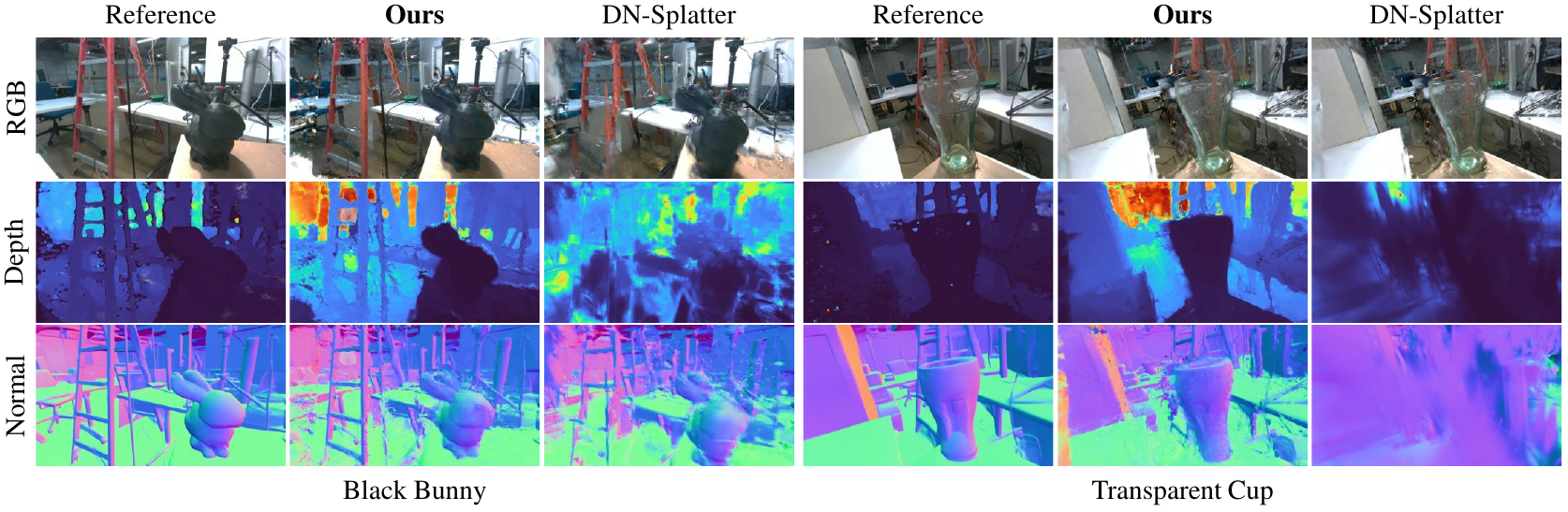}
    \vspace{-3mm}
    \caption{ \textbf{Qualitative comparisons on novel view synthesis, depth estimation, and normal estimation under sparse observations.} The comparison presents results from scenes with two challenging objects: a black bunny and a transparent Coca-Cola cup. Comparisons are made between (i) the reference (ground truth RGB images, depth images from a RealSense camera, and normal estimates generated by the DSINE monocular normal foundation model \cite{Dsine}), (ii) the proposed \textbf{\acronym} framework, and (iii) the DN-Splatter approach. Using sparse observations—9 views and 10 tactile contacts—\textbf{\acronym} achieves higher image fidelity, more precise depth, and normal estimations compared to DN-Splatter \cite{turkulainen2024dn}, which relies on 9 views.}
    \label{fig:workflow}
    \end{center}
    \vspace{-7mm}
\end{figure*}

\subsection{Experiment Setup}
\subsubsection{Robot and Sensor}
Our experiments are conducted using a GelSight Mini tactile sensor for acquiring tactile signal $\bs{T}_i$, an Intel RealSense D405 for acquiring color $\bs{C}_i$ and depth $\bs{D}_i$ images and a 6 DOF UFactory xArm 6 cobot with 0.1-millimeter repeatability. The camera and tactile sensor are mounted to the robot's end-effector with a 3D printed mount, so we know the dimensions and can easily calculate accurate transformations between each sensor and the end-effector, and therefore, the robot base, which also serves as the origin of our world frame.

\subsubsection{Data Collection and Challenging Objects}
We conduct real-world robot experiments and collect data from surrounding scenes featuring four challenging objects, comparable to the tactile baseline \cite{swann2024touch} in quantity and difficulty. The challenging objects are categorized into (1) 3D-printed transparent, reflective, and dark objects\footnote{All the files that are needed for 3D printing these objects and the sensor mount can be found in our GitHub.} and (2) non-3D-printed objects. All the 3D-printed objects are printed with a high-precision Formlabs resin printer \cite{formlabs_2024} before they are painted to achieve high transparency, reflection, or darkness.  

\subsubsection{Comparison Methods and Metrics}
We compared our method with three representative 3D Gaussian Splatting (3DGS) approaches: (i) DN-Splatter for scene and object reconstruction, (ii) GaussianObject \cite{yang2024gaussianobject} for sparse-view object reconstruction, and (iii) TouchGS\footnote{Since their method requires a large number of touches (e.g., 632), we use the results reported in \cite{swann2024touch} for comparison. We compare their 8-view against our 5-view and 152-view against our 9-view.\label{touchgs-footlabel}} for visual-tactile integration. We evaluate \textbf{scene reconstruction} using standard novel view synthesis metrics such as \textbf{PSNR} and \textbf{SSIM} \cite{kerbl20233d}. To assess \textbf{object reconstruction} quality, we calculate the \textbf{Chamfer Distance (CD)} between the reconstructed surface point clouds and ground truth point clouds, which are down-sampled from the CAD models of the 3D-printed objects.

\subsubsection{Computation}
The control of the robot and the communications between the robot, sensors, and VLM are through ROS2 \cite{ros2}. The motion planning and servoing are achieved with MoveIt2 \cite{MoveIt}. The VLM checkpoint is \texttt{gpt-4o-2024-08-06} with Structured Output. All the computation, including the inference of PartSLIP, is done on a workstation with an AMD Ryzen 9 7950X 16-Core CPU and an NVIDIA RTX 3090 GPU with 128GB RAM.

\subsubsection{Implementation Details}
The proposed Gaussian Splatting training approach is implemented using PyTorch and \texttt{gsplat} library \cite{ye2024gsplat}. To produce a well-performing visual hull, up to 5\% of the selected grid points may not be observed in the mask image. All models are trained for 15,000 iterations, and densification begins at 800 iterations. The touch patches are added at iteration 1,000 to the Gaussian scene as anchor points.

\subsection{Experiment Results}

\begin{table}
\vspace{1mm}
\begin{center}
\caption{Quantitative Comparisons of Novel View Synthesis with Varying Input Views for Scene Reconstruction}
\label{tab:sc-recon}
\begin{tabular}{ccccc}
\hline
\multirow{2}{*}{\textbf{Method}} & \multicolumn{2}{c}{\textbf{5 Views}}                 & \multicolumn{2}{c}{\textbf{9 Views}}              \\ 
                                 & \textbf{PSNR$\uparrow$}     & \textbf{SSIM$\uparrow$} & \textbf{PSNR$\uparrow$} & \textbf{SSIM$\uparrow$} \\ \hline
\textbf{DN-Splatter} \cite{turkulainen2024dn} & 13.56          & \textbf{0.58}           & 13.32                   & 0.55                    \\ 
\textbf{TouchGS\footref{touchgs-footlabel}} \cite{swann2024touch}        & 11.75          & 0.47                    & 15.51                   & \textbf{0.66}                    \\ 
\textbf{\acronym{} (Ours)}                    & \textbf{16.09} & 0.57                    & \textbf{18.83}          & 0.65                    \\ \hline
\end{tabular}
\end{center}
\vspace{-3mm}
\end{table}

\begin{table}
\begin{center}
\caption{Quantitative Comparisons of Novel View Synthesis with Varying Input Views for Object Reconstruction}
\label{tab:ob-recon}
\begin{tabular}{ccccc}
\hline
{\multirow{2}{*}{\textbf{Method}}} & \multicolumn{2}{c}{\textbf{5 Views}}             & \multicolumn{2}{c}{\textbf{9 Views}}                     \\ 
                                   & \textbf{PSNR$\uparrow$} & \textbf{SSIM$\uparrow$} & \textbf{PSNR$\uparrow$} & \textbf{SSIM$\uparrow$}        \\ \hline
\textbf{GaussianObject} \cite{yang2024gaussianobject}        & 11.38                   & \textbf{0.53}           & 12.73          & 0.57          \\ 
\textbf{DN-Splatter} \cite{turkulainen2024dn}                & 14.33                   & 0.43                    & 13.75          & 0.46          \\ 
\textbf{\acronym{} (Ours)}                                   & \textbf{18.33}          & 0.51                    & \textbf{19.84} & \textbf{0.58} \\ \hline
\end{tabular}
\end{center}
\vspace{-7mm}
\end{table}


We evaluate our method's performance in appearance and geometry representation against other representative methods. The results demonstrate that our method unifies the advantages of the other three approaches and performs better in rendering novel views and geometric representations. 

\begin{figure}
\vspace{-3mm}
\begin{center}
\includegraphics[width=0.46\textwidth]{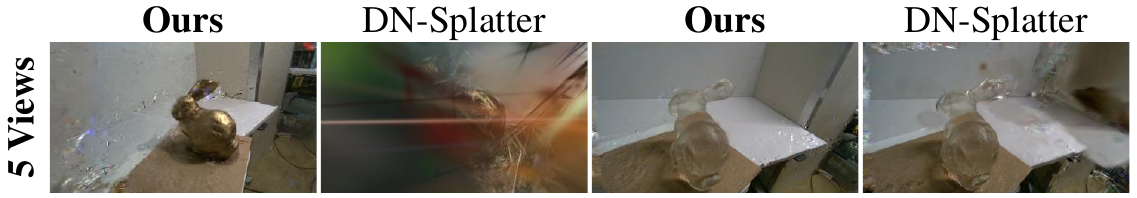}
\vspace{-3mm}
\caption{Rendering Results Using 5 Views.}
\label{fig:5-view}
\end{center}
\vspace{-7mm}
\end{figure}

Our baseline DN-splatter~\cite{turkulainen2024dn} incorporates depth and normal supervision into 3DGS training to enhance reconstruction quality with dense training views. But as shown in Fig~\ref{fig:workflow} and Fig~\ref{fig:5-view}, DN-splatter struggles with sparse views. Metrics from Table \ref{tab:sc-recon} and \ref{tab:ob-recon} further validate this observation. With insufficient views, DN-splatter's randomly initialized point cloud will likely be stuck in local minima during optimization, leading to artificial floating Gaussian points between object and scene and an incoherent representation of the target object. As a result, the extracted surface point cloud would not represent the target object well, reflected by its poor Chamfer Distance (CD) performance shown in Table \ref{tab:chamfer}.

In contrast, GaussianObject \cite{yang2024gaussianobject} initialize GS training using Visual Hull~\cite{laurentini1994visual}. However, its RGB-only supervision lacks depth and normal supervision, resulting in poor depth estimations and Gaussian gradients of the surface. Moreover, it is particularly difficult to achieve good results with RGB training alone when the target has challenging material. As Table \ref{tab:ob-recon} shows, GaussianObject performs the worst in reconstructing the four challenging objects. 

In contrast, our approach uses segmented foreground and background points as seed points, providing a better initialization of the approximate positions and coarse shapes of objects and scenes. We further enhance our approach by integrating RealSense depth data and normal priors from the foundation model for supervision. Additionally, we regularize Gaussian training through hull pruning, which removes floating Gaussian points between the object and the background. These techniques significantly reduce false occluding Gaussian points from novel perspectives, contributing to our improved rendering and geometric results. 

Besides, considering that we are more concerned with the quality of the target object than the entire scene, we use masks generated by Grounded SAM 2 \cite{ren2024grounded} to select the pixels corresponding to the target object. This allows us to calculate object-specific PSNR and SSIM metrics.

TouchGS~\cite{swann2024touch} achieves better CD with rich tactile sensing information from 632 touches by fusing touch points with the implicit surface to generate extra depth and uncertainty information. However, this approach strongly depends on the number and positioning of touches, which fails when the touch information is sparse. In contrast, our method circumvents this reliance by focusing only on empirically complex regions identified by the large language model. As shown in Table \ref{tab:chamfer}, we achieve a competitive geometrical result with just 10 touches under a sparse view scenario, as opposed to their 632 touches.

\begin{table}[]
\vspace{5pt}
\begin{center}
\caption{Chamfer Distance $\downarrow$ (mm) for Object Reconstruction}
\label{tab:chamfer}
\begin{tabular}{cccc}
\hline
\textbf{Method}                               & \textbf{\# Touches} & \textbf{5 Views}                  & \textbf{9 Views}                  \\ \hline
\textbf{DN-Splatter} \cite{turkulainen2024dn} & 0                   & 0.237                             & 0.192                             \\ 
\textbf{TouchGS} \cite{swann2024touch}        & 632                 & \textbf{0.023}                    & N/A                               \\ 
\textbf{\acronym{} (Ours)}                    & 10                  & 0.025                             & \textbf{0.022}                    \\ \hline
\end{tabular}
\end{center}
\vspace{-7mm}
\end{table}

\subsection{Ablation Study}

\subsubsection{Hull Pruning}
As mentioned in Sec. \ref{subsec: global}, hull pruning is a major modification that enables our framework. As shown in Table \ref{tab:ablation}, our framework without hull pruning suffers worse results in both scene and object reconstruction tasks. Without highly accurate depth supervision, many outliers will be generated during Gaussian Splatting field training. While the Realsense camera performs well for close-range scenes, it struggles with distant scenes and object edges. Additionally, depth estimates produced by large models like Metric3D v2 \cite{hu2024metric3d} may perform well in a single viewpoint, but they often have incorrect scaling and cannot be accurately projected into a 3D model within an entire 3D scene. Therefore, hull pruning is particularly important. It effectively prevents the Gaussian points of the target object from becoming blurred or losing edge clarity due to background interference, all while not disrupting the rendering of the surrounding scene. 

\subsubsection{Touch Strategy}
Active touch strategy is another major design in our pipeline. As shown in Table \ref{tab:ablation}, our strategy yields slightly better results, although not across the board. There are several possible reasons: (1) the number of touches is limited, and the sizes of our objects are small in the scene. Thus, it is not easy to distinguish the effectiveness of different strategies. (2) Our first module in Sec. \ref{subsec: global} gives unexpectedly outstanding precise results, leaving relatively little room for improvement for different touch strategies. 

\begin{table}[]
\vspace{5pt}
\begin{center}
\caption{Ablation Study for Hull Pruning and Touch Strategy}
\label{tab:ablation}
\begin{tabular}{cccc}
\hline
\textbf{\begin{tabular}[c]{@{}c@{}}Data\\ Description\end{tabular}}             & \textbf{Setting} & \textbf{PSNR$\uparrow$} & \textbf{\begin{tabular}[c]{@{}c@{}}CD$\downarrow$\end{tabular}} \\ \hline
\multirow{3}{*}{\textbf{\begin{tabular}[c]{@{}c@{}}Black\\ Bunny\end{tabular}}} & w/o Hull Pruning & 20.05          & 0.0305          \\ 
                                                                                & Random Touch     & 20.50          & 0.0183          \\ 
                                                                                & Active Touch     & \textbf{20.87} & \textbf{0.0176} \\ \hline
\multirow{3}{*}{\textbf{\begin{tabular}[c]{@{}c@{}}Transparent\\ Bunny\end{tabular}}} & w/o Hull Pruning & 21.46          & 0.0485          \\ 
                                                                                      & Random Touch     & 21.51          & \textbf{0.0247} \\ 
                                                                                      & Active Touch     & \textbf{21.65} & 0.0267          \\ \hline
\end{tabular}
\end{center}
\end{table}

%% file: sections/5-Conclusion.tex
\section{Conclusion, Limitation, and Future Work}
In this work, by fusing visual, tactile, and common-sense information, we propose a novel framework that significantly improves the state-of-the-art of scene and object reconstruction regarding challenging objects. Accompanying this framework, we propose a hierarchical optimization strategy designed for 3DGS that utilizes visual hull pruning and is the first to natively incorporate tactile signals into 3DGS without limiting touch numbers.

Meanwhile, we realize the limitations of our experiments and methods. Due to time constraints and the fact that some of our comparable works are close-source, we are not able to conduct a more exhaustive comparison study that includes more methods using older reconstruction methods. In addition, our touch selection strategy can use more design and experiments. Currently, its effectiveness remains marginal, and the investigation into it remains limited as the number of touches is minimal. Another limitation lies in the process of extracting point cloud and mesh from our trained Gaussian primitives. Although Gaussian points from tactile sensors are anchored as geometrical regularization, the fine-grained geometrical tactile details cannot be fully extracted from trained Gaussian scenes. This is primarily because the tiny tactile Gaussian points cannot be fully sampled during the level-set extraction approach. To handle an extensive range of multi-scaled geometrical details from scene to tactile, novel strategies need to be developed.

\textls[-1]{In the future, we plan to introduce additional constraints, such as an ideal SDF loss, to ensure that Gaussian points are optimally distributed on the surface. Currently, tactile patches are acquired through teleoperated robot control, but developing an automated, servoing-based method could significantly increase the number of touch interactions we can perform.}

%% file: sections/6-Appendix.tex
\subsection{Acquiring Tactile Patch in Real World}
There are several challenges when it comes to actually acquiring the tactile patch we need. 
\subsubsection{Object Mount}
Because the object will be touched, there needs to be a way to prevent it from moving when in contact so that the pose of the object we acquire during reconstruction can still be valid when registering the tactile patch. This demand calls for a secure way to fasten the object. We use a Peak Design camera tripod as the mount for the object so the height of the object can be freely adjusted. The tripod also has a mechanism to add dead weight so the mount will not wobble when the object is pushed by the robot. On top of the tripod, we use an Arca-Swiss plate to serve as a secure interface between the tripod and the platform that the object stands on. Arca-Swiss is a professional camera mount plate that can securely fasten a camera on a tripod. It uses a quarter-inch screw to connect to the tripod. We 3D print a mounting platform that inserts into the groove of the Arca-Swiss plate so that the platform can be securely attached to the tripod. We then glue the object of interest to a wooden plate and use two clamps to secure it on the mounting platform, as seen in Fig. \ref{fig:fig1}. Needless to say, this process is not practical and calls for investigation into how to dynamically combine tactile and visual information, even when the object of interest is moved during the process of touching. All the parts that need 3D printing can be found in our GitHub repository.
\subsubsection{Robot Control} After our algorithm gives the coordinates that need to be touched, controlling the robot to actually touch it is another challenge. We tried two methods. (1) We teleoperate the robot to touch the object given the coordinate information. (2) We calculate the normal vector of the touch coordinate and position the z-axis of the end-effector to align with that normal vector, and then use MoveIt Servo to send a twist command to drive the end-effector on the z-axis toward the object. In practice, we found no significant difference in terms of results when both methods work because the tactile patch is significantly larger than the granularity of a coordinate point, so we have a large margin of error. However, the second method often suffer from singularity issue so the success rate is lower.

%% file: root.bbl
\begin{thebibliography}{51}
\providecommand{\natexlab}[1]{#1}
\providecommand{\url}[1]{#1}
\csname url@samestyle\endcsname
\providecommand{\newblock}{\relax}
\providecommand{\bibinfo}[2]{#2}
\providecommand{\BIBentrySTDinterwordspacing}{\spaceskip=0pt\relax}
\providecommand{\BIBentryALTinterwordstretchfactor}{4}
\providecommand{\BIBentryALTinterwordspacing}{\spaceskip=\fontdimen2\font plus
\BIBentryALTinterwordstretchfactor\fontdimen3\font minus \fontdimen4\font\relax}
\providecommand{\BIBforeignlanguage}[2]{{%
\expandafter\ifx\csname l@#1\endcsname\relax
\typeout{** WARNING: IEEEtranN.bst: No hyphenation pattern has been}%
\typeout{** loaded for the language `#1'. Using the pattern for}%
\typeout{** the default language instead.}%
\else
\language=\csname l@#1\endcsname
\fi
#2}}
\providecommand{\BIBdecl}{\relax}
\BIBdecl

\bibitem[Hutmacher(2019)]{hutmacher2019there}
F.~Hutmacher, ``Why is there so much more research on vision than on any other sensory modality?'' \emph{Frontiers in psychology}, vol.~10, p. 481030, 2019.

\bibitem[Helbig and Ernst(2007)]{helbig2007optimal}
H.~B. Helbig and M.~O. Ernst, ``Optimal integration of shape information from vision and touch,'' \emph{Experimental brain research}, vol. 179, pp. 595--606, 2007.

\bibitem[Allen(1984)]{allen1984surface}
P.~Allen, ``Surface descriptions from vision and touch,'' in \emph{Proceedings. 1984 IEEE International Conference on Robotics and Automation}, vol.~1, 1984, pp. 394--397.

\bibitem[Navarro-Guerrero et~al.(2023)Navarro-Guerrero, Toprak, Josifovski, and Jamone]{navarro2023visuo}
N.~Navarro-Guerrero, S.~Toprak, J.~Josifovski, and L.~Jamone, ``Visuo-haptic object perception for robots: an overview,'' \emph{Autonomous Robots}, vol.~47, no.~4, pp. 377--403, 2023.

\bibitem[Kerbl et~al.(2023)Kerbl, Kopanas, Leimk{\"u}hler, and Drettakis]{kerbl20233d}
B.~Kerbl, G.~Kopanas, T.~Leimk{\"u}hler, and G.~Drettakis, ``3d gaussian splatting for real-time radiance field rendering,'' \emph{ACM Transactions on Graphics (TOG)}, vol.~42, no.~4, pp. 1--14, 2023.

\bibitem[Yu et~al.(2024)Yu, Sattler, and Geiger]{yu2024gaussian}
Z.~Yu, T.~Sattler, and A.~Geiger, ``Gaussian opacity fields: Efficient and compact surface reconstruction in unbounded scenes,'' \emph{arXiv preprint arXiv:2404.10772}, 2024.

\bibitem[Wang et~al.(2024)Wang, Leroy, Cabon, Chidlovskii, and Revaud]{wang2024dust3r}
S.~Wang, V.~Leroy, Y.~Cabon, B.~Chidlovskii, and J.~Revaud, ``Dust3r: Geometric 3d vision made easy,'' in \emph{Proceedings of the IEEE/CVF Conference on Computer Vision and Pattern Recognition}, 2024, pp. 20\,697--20\,709.

\bibitem[Yang et~al.(2024)Yang, Li, Fang, Liang, Xie, Zhang, Shen, and Tian]{yang2024gaussianobject}
C.~Yang, S.~Li, J.~Fang, R.~Liang, L.~Xie, X.~Zhang, W.~Shen, and Q.~Tian, ``Gaussianobject: Just taking four images to get a high-quality 3d object with gaussian splatting,'' \emph{arXiv preprint arXiv:2402.10259}, 2024.

\bibitem[Cai et~al.(2024)Cai, Qiu, Li, and Ren]{cai2024neuralto}
Y.~Cai, J.~Qiu, Z.~Li, and B.~Ren, ``Neuralto: Neural reconstruction and view synthesis of translucent objects,'' \emph{ACM Transactions on Graphics (TOG)}, vol.~43, no.~4, pp. 1--14, 2024.

\bibitem[Liu et~al.(2023{\natexlab{a}})Liu, Wang, Lin, Long, Wang, Liu, Komura, and Wang]{liu2023nero}
Y.~Liu, P.~Wang, C.~Lin, X.~Long, J.~Wang, L.~Liu, T.~Komura, and W.~Wang, ``Nero: Neural geometry and brdf reconstruction of reflective objects from multiview images,'' \emph{ACM Transactions on Graphics (TOG)}, vol.~42, no.~4, pp. 1--22, 2023.

\bibitem[Swann et~al.(2024)Swann, Strong, Do, Camps, Schwager, and Kennedy~III]{swann2024touch}
A.~Swann, M.~Strong, W.~K. Do, G.~S. Camps, M.~Schwager, and M.~Kennedy~III, ``Touch-gs: Visual-tactile supervised 3d gaussian splatting,'' \emph{arXiv preprint arXiv:2403.09875}, 2024.

\bibitem[Comi et~al.(2024)Comi, Lin, Church, Tonioni, Aitchison, and Lepora]{comi2024touchsdf}
M.~Comi, Y.~Lin, A.~Church, A.~Tonioni, L.~Aitchison, and N.~F. Lepora, ``Touchsdf: A deepsdf approach for 3d shape reconstruction using vision-based tactile sensing,'' \emph{IEEE Robotics and Automation Letters}, 2024.

\bibitem[Park et~al.(2019)Park, Florence, Straub, Newcombe, and Lovegrove]{park2019deepsdf}
J.~J. Park, P.~Florence, J.~Straub, R.~Newcombe, and S.~Lovegrove, ``Deepsdf: Learning continuous signed distance functions for shape representation,'' in \emph{Proceedings of the IEEE/CVF conference on computer vision and pattern recognition}, 2019, pp. 165--174.

\bibitem[Yuan et~al.(2017)Yuan, Dong, and Adelson]{yuan2017gelsight}
W.~Yuan, S.~Dong, and E.~H. Adelson, ``Gelsight: High-resolution robot tactile sensors for estimating geometry and force,'' \emph{Sensors}, vol.~17, no.~12, p. 2762, 2017.

\bibitem[Wang et~al.(2021)Wang, She, Romero, and Adelson]{wang2021gelsight}
S.~Wang, Y.~She, B.~Romero, and E.~Adelson, ``Gelsight wedge: Measuring high-resolution 3d contact geometry with a compact robot finger,'' in \emph{2021 IEEE International Conference on Robotics and Automation (ICRA)}, 2021, pp. 6468--6475.

\bibitem[Shahidzadeh et~al.(2024)Shahidzadeh, Yoo, Mantripragada, Singh, Ferm{\"u}ller, and Aloimonos]{shahidzadeh2024actexplore}
A.-H. Shahidzadeh, S.~J. Yoo, P.~Mantripragada, C.~D. Singh, C.~Ferm{\"u}ller, and Y.~Aloimonos, ``Actexplore: Active tactile exploration on unknown objects,'' in \emph{2024 IEEE International Conference on Robotics and Automation (ICRA)}, 2024, pp. 3411--3418.

\bibitem[Calli et~al.(2015)Calli, Singh, Walsman, Srinivasa, Abbeel, and Dollar]{calli2015ycb}
B.~Calli, A.~Singh, A.~Walsman, S.~Srinivasa, P.~Abbeel, and A.~M. Dollar, ``The ycb object and model set: Towards common benchmarks for manipulation research,'' in \emph{2015 international conference on advanced robotics (ICAR)}, 2015, pp. 510--517.

\bibitem[Suresh et~al.(2022)Suresh, Si, Mangelson, Yuan, and Kaess]{suresh2022shapemap}
S.~Suresh, Z.~Si, J.~G. Mangelson, W.~Yuan, and M.~Kaess, ``Shapemap 3-d: Efficient shape mapping through dense touch and vision,'' in \emph{2022 International Conference on Robotics and Automation (ICRA)}, 2022, pp. 7073--7080.

\bibitem[Bae and Davison(2024{\natexlab{a}})]{bae2024rethinking}
G.~Bae and A.~J. Davison, ``Rethinking inductive biases for surface normal estimation,'' in \emph{Proceedings of the IEEE/CVF Conference on Computer Vision and Pattern Recognition}, 2024, pp. 9535--9545.

\bibitem[Turkulainen et~al.(2024)Turkulainen, Ren, Melekhov, Seiskari, Rahtu, and Kannala]{turkulainen2024dn}
M.~Turkulainen, X.~Ren, I.~Melekhov, O.~Seiskari, E.~Rahtu, and J.~Kannala, ``Dn-splatter: Depth and normal priors for gaussian splatting and meshing,'' \emph{arXiv preprint arXiv:2403.17822}, 2024.

\bibitem[Mazur et~al.(2024)Mazur, Bae, and Davison]{asuperprimitive}
K.~Mazur, G.~Bae, and A.~Davison, ``{SuperPrimitive}: Scene reconstruction at a primitive level,'' in \emph{IEEE/CVF Conference on Computer Vision and Pattern Recognition (CVPR)}, 2024.

\bibitem[Cao and Taketomi(2024)]{cao2024supernormal}
X.~Cao and T.~Taketomi, ``Supernormal: Neural surface reconstruction via multi-view normal integration,'' in \emph{Proceedings of the IEEE/CVF Conference on Computer Vision and Pattern Recognition}, 2024, pp. 20\,581--20\,590.

\bibitem[Goldberg and Bajcsy(1984)]{Goldberg1984ActiveTA}
K.~Goldberg and R.~Bajcsy, ``Active touch and robot perception,'' in \emph{Cognition and Brain Theory}, 1984.

\bibitem[Tomo et~al.(2018)Tomo, Schmitz, Wong, Kristanto, Somlor, Hwang, Jamone, and Sugano]{uskin}
T.~P. Tomo, A.~Schmitz, W.~K. Wong, H.~Kristanto, S.~Somlor, J.~Hwang, L.~Jamone, and S.~Sugano, ``Covering a robot fingertip with uskin: A soft electronic skin with distributed 3-axis force sensitive elements for robot hands,'' \emph{IEEE Robotics and Automation Letters}, vol.~3, no.~1, pp. 124--131, 2018.

\bibitem[Corcodel et~al.(2020)Corcodel, Jain, and van Baar]{corcodel2020interactive}
R.~Corcodel, S.~Jain, and J.~van Baar, ``Interactive tactile perception for classification of novel object instances,'' in \emph{2020 IEEE/RSJ International Conference on Intelligent Robots and Systems (IROS)}, 2020, pp. 9861--9868.

\bibitem[Schmitz et~al.(2014)Schmitz, Bansho, Noda, Iwata, Ogata, and Sugano]{2014Schmitz}
A.~Schmitz, Y.~Bansho, K.~Noda, H.~Iwata, T.~Ogata, and S.~Sugano, ``Tactile object recognition using deep learning and dropout,'' in \emph{2014 IEEE-RAS International Conference on Humanoid Robots}, 2014, pp. 1044--1050.

\bibitem[Burns et~al.(2022)Burns, Xiang, Lee, Jackel, Song, and Isler]{Burns2022}
A.~Burns, S.~Xiang, D.~Lee, L.~Jackel, S.~Song, and V.~Isler, ``Look and listen: A multi-sensory pouring network and dataset for granular media from human demonstrations,'' in \emph{2022 International Conference on Robotics and Automation (ICRA)}, 2022, pp. 2519--2524.

\bibitem[Kaboli et~al.(2016)Kaboli, Yao, and Cheng]{Kaboli2016}
M.~Kaboli, K.~Yao, and G.~Cheng, ``Tactile-based manipulation of deformable objects with dynamic center of mass,'' in \emph{2016 IEEE-RAS 16th International Conference on Humanoid Robots (Humanoids)}, 2016, pp. 752--757.

\bibitem[She et~al.(2021)She, Wang, Dong, Sunil, Rodriguez, and Adelson]{she2021cable}
Y.~She, S.~Wang, S.~Dong, N.~Sunil, A.~Rodriguez, and E.~Adelson, ``Cable manipulation with a tactile-reactive gripper,'' \emph{The International Journal of Robotics Research}, vol.~40, no. 12-14, pp. 1385--1401, 2021.

\bibitem[Dong and Rodriguez(2019)]{2019DOng}
S.~Dong and A.~Rodriguez, ``Tactile-based insertion for dense box-packing,'' in \emph{2019 IEEE/RSJ International Conference on Intelligent Robots and Systems (IROS)}, 2019, pp. 7953--7960.

\bibitem[Dong et~al.(2021)Dong, Jha, Romeres, Kim, Nikovski, and Rodriguez]{Dong2021}
S.~Dong, D.~K. Jha, D.~Romeres, S.~Kim, D.~Nikovski, and A.~Rodriguez, ``Tactile-rl for insertion: Generalization to objects of unknown geometry,'' in \emph{2021 IEEE International Conference on Robotics and Automation (ICRA)}, 2021, pp. 6437--6443.

\bibitem[Yi et~al.(2016)Yi, Calandra, Veiga, van Hoof, Hermans, Zhang, and Peters]{Yi2016}
Z.~Yi, R.~Calandra, F.~Veiga, H.~van Hoof, T.~Hermans, Y.~Zhang, and J.~Peters, ``Active tactile object exploration with gaussian processes,'' in \emph{2016 IEEE/RSJ International Conference on Intelligent Robots and Systems (IROS)}, 2016, pp. 4925--4930.

\bibitem[Jamali et~al.(2016)Jamali, Ciliberto, Rosasco, and Natale]{Jamali2016}
N.~Jamali, C.~Ciliberto, L.~Rosasco, and L.~Natale, ``Active perception: Building objects' models using tactile exploration,'' in \emph{2016 IEEE-RAS 16th International Conference on Humanoid Robots (Humanoids)}, 2016, pp. 179--185.

\bibitem[Matsubara and Shibata(2017)]{MATSUBARA2017314}
T.~Matsubara and K.~Shibata, ``Active tactile exploration with uncertainty and travel cost for fast shape estimation of unknown objects,'' \emph{Robotics and Autonomous Systems}, vol.~91, pp. 314--326, 2017.

\bibitem[Smith et~al.(2020)Smith, Calandra, Romero, Gkioxari, Meger, Malik, and Drozdzal]{smith20203d}
E.~Smith, R.~Calandra, A.~Romero, G.~Gkioxari, D.~Meger, J.~Malik, and M.~Drozdzal, ``3d shape reconstruction from vision and touch,'' \emph{Advances in Neural Information Processing Systems}, vol.~33, pp. 14\,193--14\,206, 2020.

\bibitem[Smith et~al.(2021)Smith, Meger, Pineda, Calandra, Malik, Romero~Soriano, and Drozdzal]{smith2021active}
E.~Smith, D.~Meger, L.~Pineda, R.~Calandra, J.~Malik, A.~Romero~Soriano, and M.~Drozdzal, ``Active 3d shape reconstruction from vision and touch,'' \emph{Advances in Neural Information Processing Systems}, vol.~34, pp. 16\,064--16\,078, 2021.

\bibitem[Bj{\"o}rkman et~al.(2013)Bj{\"o}rkman, Bekiroglu, H{\"o}gman, and Kragic]{bjorkman2013enhancing}
M.~Bj{\"o}rkman, Y.~Bekiroglu, V.~H{\"o}gman, and D.~Kragic, ``Enhancing visual perception of shape through tactile glances,'' in \emph{2013 IEEE/RSJ International Conference on Intelligent Robots and Systems (IROS)}, 2013, pp. 3180--3186.

\bibitem[Wang et~al.(2018)Wang, Wu, Sun, Yuan, Freeman, Tenenbaum, and Adelson]{wang20183d}
S.~Wang, J.~Wu, X.~Sun, W.~Yuan, W.~T. Freeman, J.~B. Tenenbaum, and E.~H. Adelson, ``3d shape perception from monocular vision, touch, and shape priors,'' in \emph{2018 IEEE/RSJ International Conference on Intelligent Robots and Systems (IROS)}, 2018, pp. 1606--1613.

\bibitem[Williams and Fitzgibbon(2006)]{williams2006gaussian}
O.~Williams and A.~Fitzgibbon, ``Gaussian process implicit surfaces,'' in \emph{Gaussian Processes in Practice}, 2006.

\bibitem[Hu et~al.(2024)Hu, Yin, Zhang, Cai, Long, Chen, Wang, Yu, Shen, and Shen]{hu2024metric3d}
M.~Hu, W.~Yin, C.~Zhang, Z.~Cai, X.~Long, H.~Chen, K.~Wang, G.~Yu, C.~Shen, and S.~Shen, ``Metric3d v2: A versatile monocular geometric foundation model for zero-shot metric depth and surface normal estimation,'' \emph{arXiv preprint arXiv:2404.15506}, 2024.

\bibitem[Laurentini(1994)]{laurentini1994visual}
A.~Laurentini, ``The visual hull concept for silhouette-based image understanding,'' \emph{IEEE Transactions on pattern analysis and machine intelligence}, vol.~16, no.~2, pp. 150--162, 1994.

\bibitem[Ren et~al.(2024)Ren, Liu, Zeng, Lin, Li, Cao, Chen, Huang, Chen, Yan, et~al.]{ren2024grounded}
T.~Ren, S.~Liu, A.~Zeng, J.~Lin, K.~Li, H.~Cao, J.~Chen, X.~Huang, Y.~Chen, F.~Yan \emph{et~al.}, ``Grounded sam: Assembling open-world models for diverse visual tasks,'' \emph{arXiv preprint arXiv:2401.14159}, 2024.

\bibitem[Bae and Davison(2024{\natexlab{b}})]{Dsine}
\BIBentryALTinterwordspacing
G.~Bae and A.~J. Davison, ``Rethinking inductive biases for surface normal estimation,'' 2024. [Online]. Available: \url{https://arxiv.org/abs/2403.00712}
\BIBentrySTDinterwordspacing

\bibitem[Ester et~al.(1996)Ester, Kriegel, Sander, and Xu]{DBSCAN}
M.~Ester, H.-P. Kriegel, J.~Sander, and X.~Xu, ``A density-based algorithm for discovering clusters in large spatial databases with noise,'' in \emph{Proceedings of the Second International Conference on Knowledge Discovery and Data Mining}, ser. KDD'96.\hskip 1em plus 0.5em minus 0.4em\relax AAAI Press, 1996, p. 226–231.

\bibitem[OpenAI et~al.(2023)]{openai2023gpt}
OpenAI \emph{et~al.}, ``Gpt-4 technical report,'' \emph{arXiv preprint arXiv:2303.08774}, 2023.

\bibitem[Liu et~al.(2023{\natexlab{b}})Liu, Zhu, Cai, Han, Ling, Porikli, and Su]{liu2023partslip}
M.~Liu, Y.~Zhu, H.~Cai, S.~Han, Z.~Ling, F.~Porikli, and H.~Su, ``Partslip: Low-shot part segmentation for 3d point clouds via pretrained image-language models,'' in \emph{Proceedings of the IEEE/CVF conference on computer vision and pattern recognition}, 2023, pp. 21\,736--21\,746.

\bibitem[Woodham(1992)]{Woodham1992}
R.~Woodham, ``Photometric method for determining surface orientation from multiple images,'' \emph{Optical Engineering}, vol.~19, 1992.

\bibitem[Formlabs(2024)]{formlabs_2024}
Formlabs, ``High resolution sla and sls 3d printers for professionals,'' \url{https://formlabs.com/}, 2024, accessed: 2024-09-12.

\bibitem[Macenski et~al.(2022)Macenski, Foote, Gerkey, Lalancette, and Woodall]{ros2}
S.~Macenski, T.~Foote, B.~Gerkey, C.~Lalancette, and W.~Woodall, ``Robot operating system 2: Design, architecture, and uses in the wild,'' \emph{Science Robotics}, vol.~7, no.~66, 2022.

\bibitem[Coleman et~al.(2014)Coleman, Sucan, Chitta, and Correll]{MoveIt}
D.~Coleman, I.~A. Sucan, S.~Chitta, and N.~Correll, ``Reducing the barrier to entry of complex robotic software: a moveit! case study,'' \emph{arXiv preprint arXiv:1404.3785}, 2014.

\bibitem[Ye et~al.(2024)Ye, Li, Kerr, Turkulainen, Yi, Pan, Seiskari, Ye, Hu, Tancik, and Kanazawa]{ye2024gsplat}
V.~Ye, R.~Li, J.~Kerr, M.~Turkulainen, B.~Yi, Z.~Pan, O.~Seiskari, J.~Ye, J.~Hu, M.~Tancik, and A.~Kanazawa, ``gsplat: An open-source library for gaussian splatting,'' \emph{arXiv preprint arXiv:2409.06765}, 2024.

\end{thebibliography}
